\documentclass[letterpaper]{article} 
\usepackage{aaai25}  
\usepackage{times}  
\usepackage{helvet}  
\usepackage{courier}  
\usepackage[hyphens]{url}  
\usepackage{graphicx} 
\usepackage{natbib}  
\usepackage{caption} 
\usepackage{amssymb}
\usepackage{multirow}
\usepackage{booktabs}
\usepackage{amsmath}
\usepackage{pifont}

\title{Dynamic Contrastive Knowledge Distillation for Efficient Image Restoration}
\author{
    Yunshuai Zhou\textsuperscript{\rm 1}\equalcontrib,
    Junbo Qiao\textsuperscript{\rm 1}\equalcontrib,
    Jincheng Liao\textsuperscript{\rm 1},
    Wei Li\textsuperscript{\rm 2},
    Simiao Li\textsuperscript{\rm 2},
    Jiao Xie\textsuperscript{\rm 1},
    Yunhang Shen\textsuperscript{\rm 3},\\
    Jie Hu\textsuperscript{\rm 2},
    Shaohui Lin\textsuperscript{\rm 1,\rm 4}\footnote{Corresponding author (e-mail: shlin@cs.ecnu.edu.cn)}
}
\affiliations{
    \textsuperscript{\rm 1}East China Normal University, Shanghai, China\\
    \textsuperscript{\rm 2}Huawei Noah's Ark Lab, China\\
    \textsuperscript{\rm 3}Xiamen University, China\\
    \textsuperscript{\rm 4}Key Laboratory of Advanced Theory and Application in Statistics and Data Science - MOE, China\\
    
}

\nocopyright

\begin{document}

\maketitle

\begin{abstract}
Knowledge distillation (KD) is a valuable yet challenging approach that enhances a compact student network by learning from a high-performance but cumbersome teacher model. However, previous KD methods for image restoration overlook the state of the student during the distillation, adopting a fixed solution space that limits the capability of KD. Additionally, relying solely on L1-type loss struggles to leverage the distribution information of images. In this work, we propose a novel dynamic contrastive knowledge distillation~(DCKD) framework for image restoration. Specifically, we introduce dynamic contrastive regularization to perceive the student's learning state and dynamically adjust the distilled solution space using contrastive learning. Additionally, we also propose a distribution mapping module to extract and align the pixel-level category distribution of the teacher and student models. Note that the proposed DCKD is a structure-agnostic distillation framework, which can adapt to different backbones and can be combined with methods that optimize upper-bound constraints to further enhance model performance. Extensive experiments demonstrate that DCKD significantly outperforms the state-of-the-art KD methods across various image restoration tasks and backbones. Our codes are available at https://github.com/super-SSS/DCKD.
\end{abstract}


\section{Introduction}
Image restoration aims to recover high-quality images from low-quality ones degraded by processes such as sub-sampling, blurring, and rain streaks. It is a highly challenging ill-posed inverse problem since crucial content information is missing during degradation. Recently, convolutional neural networks~(CNNs)~\cite{dong2015image, lim2017enhanced, zhang2018image} and Transformers~\cite{liang2021swinir, chen2021pre, wang2022uformer, zamir2022restormer} have been extensively investigated for designing various models, achieving remarkable success in image restoration. However, these models demand high resources and exhibit inefficiencies, making deployment on resource-constrained devices challenging. To facilitate their real-world applications, there is a growing research focus on compressing image restoration models.

\begin{figure}[t]
    \centering
    \includegraphics[width=\columnwidth]{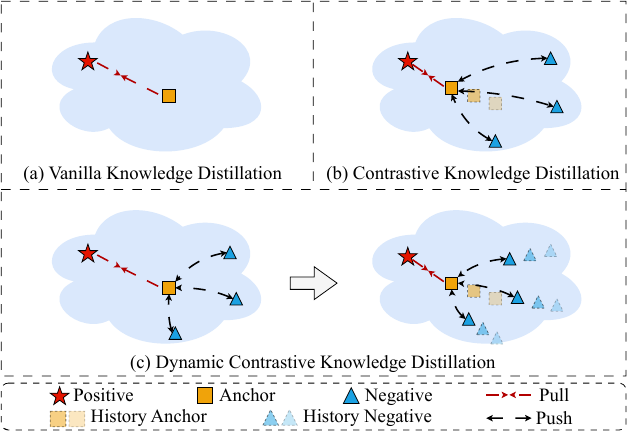}
    \caption{Difference between our DCKD and existing KD methods. (a) The Vanilla KD method overlooks the information from negative images as a lower bound. (b) Existing contrastive KD methods adopt a fixed lower bound that limits the capability of KD. (c) Our DCKD introduces dynamic contrastive regularization to perceive the student’s learning state and dynamically adjust the distilled solution space.}
    \label{fig:compare}
\end{figure}

Knowledge distillation~(KD) is an effective model compression method that transfers knowledge from a cumbersome teacher model to a lightweight student model. This process allows the student model to inherit the capabilities of the teacher model, resulting in significant performance improvements while reducing computational and storage requirements. KD has gained broad recognition for its excellent performance and broad applicability. It also can be combined with other model compression techniques, such as quantization~\cite{du2021anchor, ayazoglu2021extremely, hong2022cadyq}, pruning~\cite{fan2020neural, wang2021exploring, oh2022attentive}, compact architecture design~\cite{ahn2018fast, zhang2022efficient, chen2022simple}, and neural architecture search~(NAS)~\cite{gou2020clearer, kim2021searching, he2022spectrum}, to enhance the compactness of student models further.

Since KD has been well-established in natural language processing~\cite{hahn2019self, sanh2019distilbert, fu2021lrc} and high-level vision tasks~\cite{touvron2021training, lin2022knowledge, chen2022dearkd}, researchers have been investigating KD for image restoration methods~\cite{he2020fakd, lee2020learning, wang2021towards, zhang2023data, li2024knowledge, jiang2024mtkd, zhang2024distilling}. However, these methods adopt a fixed solution space, which limits their adaptability to the evolving state of the student model during the distillation process. As illustrated in Fig.~\ref{fig:compare} (a), the vanilla distillation method~\cite{hinton2015distilling} only constrains the upper bound of the solution space. Although existing works~\cite{zhang2023data, jiang2024mtkd} explore more effective and diverse upper bounds, the lack of constraints on the lower bound of the output image increases the difficulty of optimizing the solution space. This often generates low-quality images with artifacts, color distortion, and blurring. As illustrated in Fig.~\ref{fig:compare} (b), CSD~\cite{wang2021towards} introduces contrastive learning to design lower-bound constraints, significantly enhancing the transfer of knowledge from the teacher. However, in the later stages of training, the student anchor moves far from the lower bound, leading to a diminished constraint effect.

To address this problem, we propose a novel dynamic contrastive knowledge distillation framework named DCKD. Specifically, we first propose the dynamic contrastive regularization, which generates dynamic lower bound constraints. In addition, we also propose a distribution mapping module~(DMM) to extract and align the pixel-level category distribution between the output of the teacher and student. Compared with previous image restoration distillation methods that primarily relied on L1 loss, DMM successfully introduces category distribution information distillation into low-level vision tasks. DCKD not only adapts to various backbones but also can be combined with methods~\cite{zhang2023data, li2024knowledge, jiang2024mtkd} that improve the upper bound of the solution space to further enhance distillation performance. We validate the effectiveness of DCKD across multiple image restoration tasks, including image super-resolution, deblurring, and deraining.

Overall, our main contributions are summarized as:
\begin{itemize}
\item We propose a dynamic contrastive distillation framework~(DCKD), which can perceive the student's learning state and dynamically optimize the lower bound of the solution space.

\item We introduce a distribution mapping module to leverage category distribution information distillation, which has been significantly ignored in previous KD works for image restoration.

\item Extensive experiments across various image restoration tasks demonstrate that the proposed DCKD framework significantly outperforms previous methods.
\end{itemize}

\section{Related Work}

\subsection{Image Restoration}
Since the pioneering works SRCNN~\cite{dong2015image} and DnCNN~\cite{zhang2017beyond} are firstly to employ CNNs for image restoration, various works~\cite{lim2017enhanced, nah2017deep, lefkimmiatis2017non, li2018cascaded, ren2019progressive, chen2022simple} have been proposed to improve the performance by increasing the parameters. Recently, Transformer-based methods~\cite{chen2021pre, liang2021swinir, chen2022cross, chen2023activating} have leveraged self-attention mechanisms to capture long-range dependencies, leading to significant performance improvements in image restoration. To reduce computational overhead, SAFMN~\cite{sun2023spatially} enhances model efficiency by utilizing spatially adaptive feature pyramid attention maps. Restormer~\cite{zamir2022restormer} designs channel self-attention, which is more efficient than spatial self-attention. Although lightweight designs of CNN and Transformer architectures significantly reduce computational overhead, they still face challenges regarding direct deployment on resource-constrained platforms.

\subsection{Knowledge Distillation for Image Restoration}
Knowledge distillation aims to significantly reduce deployment costs while improving the performance of student models by emulating the behavior of teacher models~\cite{hinton2015distilling, lee2020learning, gou2021knowledge}. In recent years, numerous works have focused on knowledge distillation for image restoration. He~\textit{et al.} proposed FAKD to align the spatial affinity matrix of the feature maps between the teacher and student models~\cite{he2020fakd}. To alleviate the semantic differences between features of the teacher and student, Li~\textit{et al.} proposed MiPKD, which achieves feature and stochastic network block mixture in latent space~\cite{li2024knowledge}. Jiang~\textit{et al.} proposed MTKD that designs a composite output from multiple teachers to provide the student with a more robust teacher model~\cite{jiang2024mtkd}. However, these KD methods primarily focus on improving the upper bound of the model's solution space, without leveraging the distribution information of the images.

\subsection{Contrastive Learning for Knowledge Distillation}
Recently, many researchers have explored the combination of contrastive learning and knowledge distillation to construct a comprehensive solution space~\cite{tian2019contrastive, xu2020knowledge, yang2023online}. Wang~\textit{et al.} introduced contrastive learning in knowledge distillation, utilizing other images within the same dataset to construct a lower bound for the solution space~\cite{wang2021improving}. Similarly, CSD~\cite{wang2021towards} proposed a contrastive self-distillation method, utilizing different images within the batch to provide lower bound constraints. Luo~\textit{et al.} enriched the lower bound constraints in the image deraining task by altering the direction of falling raindrops for the student model~\cite{luo2023local}. However, these methods employ a fixed solution space, which leads to a weakening of the lower-bound constraint when the student anchor moves away from the lower bound in the later stages of training. Different from these methods, DCKD proposes a dynamic lower-bound constraint that progressively narrows the solution space, and can also be combined with enhanced upper-bound approaches. Additionally, DMM is proposed to extract and align the pixel-level category distribution information between the teacher and student networks.

\begin{figure*}[t]
    \centering
    \includegraphics[width=0.99\textwidth]{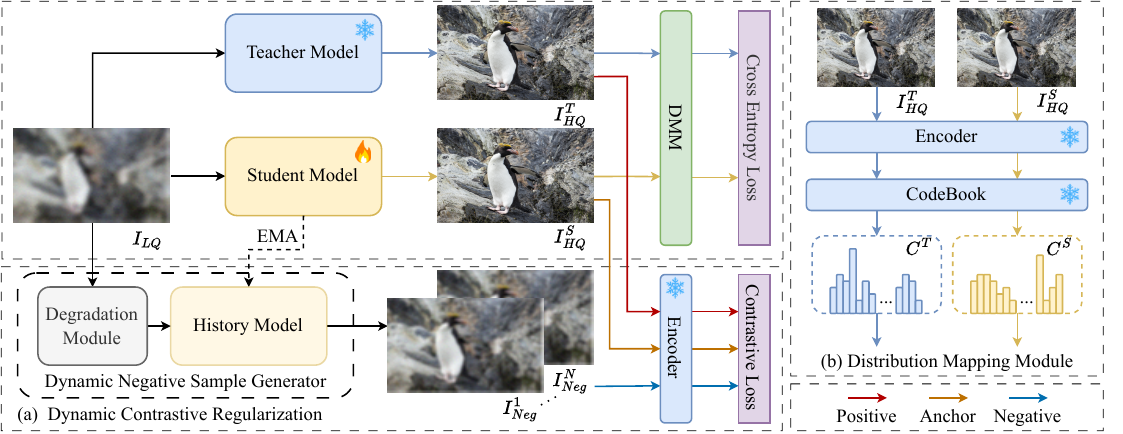}
    \caption{Illustration of the proposed Dynamic Contrastive Knowledge Distillation framework. Our DCKD consists of two parts: (a) Dynamic Contrastive Regularization~(DCR), and (b) Distribution Mapping Module~(DMM).}
    \label{fig:main}
\end{figure*}

\section{Methodology}

\subsection{Preliminaries}
Given a low-quality image $I_{LQ}$ as input, the image restoration (IR) model $\mathcal{F}(\cdot)$ generates the corresponding high-quality image $I_{HQ}$, which can be formulated as:
\begin{equation}
    I_{HQ} = \mathcal{F}(I_{LQ}; \theta),
\end{equation}
where $\theta$ represents the model parameters. The IR model $\mathcal{F}(\cdot; \theta)$ is typically optimized using the L1 norm reconstruction loss, which is defined as:
\begin{equation}
    \mathcal{L}_{rec} = \| I_{HQ} - I_{GT} \|_1,
    \label{equ:rec}
\end{equation}
where the $I_{GT}$ is the ground-truth image. The vanilla knowledge distillation method adds a KD loss to minimize the difference between the student model and the teacher model:
\begin{equation}
    \mathcal{L}_{kd} = \| \mathcal{F}_{S}(I_{LQ}; \theta_{s}) - \mathcal{F}_{T}(I_{LQ}; \theta_{t}) \|_1,
    \label{equ:kd}
\end{equation}
where $\mathcal{F}_{S}(I_{LQ};\theta_{s})$ and $\mathcal{F}_{T}(I_{LQ};\theta_{t})$ represent the outputs of student and teacher models, respectively.

\subsection{Dynamic Contrastive Regularization}
Previous KD methods~\cite{hinton2015distilling, wang2021towards} employed a fixed solution space, which causes the lower-bound constraints on the student model to weaken in the later stages of training. To address this problem, we propose dynamic contrastive regularization~(DCR), which dynamically adjusts the solution space based on the student model's state.

As shown in Fig.~\ref{fig:main} (a), we first feed the input image $I_{LQ}$ into the dynamic negative sample generator, which consists of the degradation module and the history model. The degradation module applies random degradation operations on $I_{LQ}$, generating $N$ different degraded images $I_{dirty}$. The history model $\mathcal{F}_{S}^{his}$ then reconstructs these degraded images based on the historical state of the student model, producing $N$ different negative images $I_{Neg}$ as the lower-bound of the solution space:
\begin{equation}
    I_{Neg}^{1}, \cdots, I_{Neg}^{N} = \mathcal{F}_{S}^{his}(\mathbb{D}_{1}(I_{LQ}), \cdots, \mathbb{D}_{N}(I_{LQ})),
\end{equation}
where $\mathbb{D}_{N}(\cdot)$ and $I_{Neg}^{N}$ represent the $N$-th type of random degradation and the corresponding negative image, respectively.

For the upper-bound of the solution space, we not only rely on ground-truth $I_{GT}$ to optimize the student output, as described in Eq.~\ref{equ:rec} but also consider the output of teacher model $I_{HQ}^{T}$ as the positive image $I_{Pos}$. At this point, we have defined the upper-bound and the dynamic lower-bound of the model, allowing us to construct a novel dynamic contrastive loss. We employ the pre-trained VQGAN~\cite{esser2021taming} as the feature encoder. The dynamic contrastive loss is formulated as follows:
\begin{equation}
    \mathcal{L}_{dcl} = \sum_{i=1}^{L}\lambda_{i}\frac{||f_{i}^{Anc}-f_{i}^{Pos}||_1}{\sum_{j=1}^{N}||f_{i}^{Anc}-f_{i,j}^{Neg}||_1},
    \label{equ:dcl}
\end{equation}
where $f^{Anc}$, $f^{Pos}$, and $f^{Neg}$ represent the features extracted from the output of the student model, positive image and negative images, respectively. $i$ denotes the $i$-th layer of Encoder. $\lambda_{i}$ is the balancing weight for the $i$-th layer.

To better capture the state of the student model, we introduce exponential moving averages (EMA) to update the history model:
\begin{equation}
    \theta_{his} = \alpha\theta_{his}+(1-\alpha)\theta_{stu},~~\mathrm{s.t.}~~t\%s=0,
    \label{equ:ema}
\end{equation}
where $\theta_{his}$ and $\theta_{stu}$ represent the parameters of the history model and the current student model, respectively. $\alpha$ is the attenuation rate, $t$ is the current iteration, and $s$ denotes the update step. During training, the update step $s$ gradually increases.

\begin{figure}[t]
    \centering
    \includegraphics[width=\columnwidth]{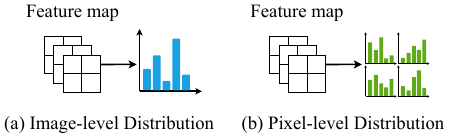}
    \caption{Illustration of the Image-level Distribution and our Pixel-level Distribution (DMM).}
    \label{fig:dmm}
\end{figure}

\subsection{Distribution Mapping Module}
Existing image restoration KD methods~\cite{li2024knowledge, jiang2024mtkd} only rely on L1-type loss to align the teacher and student models, thereby overlooking the distribution information of image content. However, high-level task KD methods~\cite{hinton2015distilling, park2019relational, huang2022knowledge} that align the entire output distribution of the teacher and student networks fail in low-level vision tasks. To address this, we design a distribution mapping module~(DMM) to extract and align pixel-wise image distribution information, which is well-suited for pixel-level image restoration tasks.

As shown in Fig.~\ref{fig:main} (b), we employ a pre-trained image encoder to extract deep features $F^{T}$ and $F^{S}$ from the output images of the teacher model $I_{HQ}^{T}$ and the student model $I_{HQ}^{S}$, respectively:
\begin{equation}
    F^{T} = \operatorname{Encoder}\left(I_{HQ}^{T}\right),
    F^{S} = \operatorname{Encoder}\left(I_{HQ}^{S}\right).
\end{equation}

We assume that high-level KD methods struggle to provide fine-grained distribution constraints, which are crucial for restoring image details. Inspired by VQGAN~\cite{esser2021taming}, we use the codebook $e$ pre-trained on ImageNet~\cite{krizhevsky2012imagenet} to obtain the pixel-wised category distribution as illustrated in Fig.~\ref{fig:dmm}. We also employ the corresponding VQGAN as the image encoder. The codebook further transforms the extracted deep features $F^{T}$ and $F^{S}$ into category distributions can be formulated as:
\begin{equation}
    C^T = \psi(\| F^{T}-e \|_{2}^{2}), C^S = \psi(\| F^{S}-e \|_{2}^{2}),
\end{equation}
where $C^T$ and $C^S$ represent the pixel-wised category distributions of the teacher and student models, respectively. $\psi$ denotes the softmax operation.

Finally, we use cross-entropy loss to align $C^T$ and $C^S$:
\begin{equation}
    \mathcal{L}_{ce} = -\sum_{i=1}^{M}C_{i}^{T}\log{C_{i}^{S}},
    \label{equ:ce}
\end{equation}
where $C_{i},i=1,2,\cdots M$ is category i. $M$ is the total number of categories.

\subsection{Overall Loss}
Following~\cite{hinton2015distilling, zhang2023data, li2024knowledge}, our DCKD also compute the reconstruction loss $\mathcal{L}_{res}$ in Eq.~\ref{equ:rec} and the vanilla distillation loss $\mathcal{L}_{kd}$ in Eq.~\ref{equ:kd}. In addition, the dynamic contrastive loss $\mathcal{L}_{dcl}$ in Eq.~\ref{equ:dcl} and the cross-entropy loss $\mathcal{L}_{ce}$ in Eq.~\ref{equ:ce} are also accumulated. The overall loss function can be expressed as:
\begin{equation}
    \mathcal{L}=\mathcal{L}_{rec}+\mathcal{L}_{kd}+\lambda_{dcl}\mathcal{L}_{dcl}+\lambda_{ce}\mathcal{L}_{ce},
    \label{equ:all}
\end{equation}
where the $\lambda_{dcl}$ and $\lambda_{ce}$ are the balancing parameters. The teacher model and the encoder are frozen during the training stage.

\section{Experiments}

\subsection{Experimental Settings}

\subsubsection{Teacher Backbones}
The proposed DCKD is evaluated on three image restoration tasks: image super-resolution, image deblurring, and image deraining. Following~\cite{li2024knowledge}, we verify the effectiveness of DCKD on Transformer-based SwinIR~\cite{liang2021swinir} and CNN-based RCAN~\cite{zhang2018image} in image super-resolution. For image deblurring, we use NAFNet~\cite{chen2022simple} and Restormer~\cite{zamir2022restormer} as the teacher backbones. For image deraining, we employ Restormer as the backbone. The configuration details of teacher and student models are presented in Tab.~\ref{tab:config}.

\begin{table}[h]
\small
\centering
\begin{tabular}{c|c|c|c|c}
\toprule
Model & Role & Channel & Block/Group & \#Params \\
\midrule
\multirow{2}{*}{SwinIR} & Teacher & 180 & 6/- & 11.9M \\
& Student & 60 & 4/- & 1.2M \\
\midrule
\multirow{2}{*}{RCAN} & Teacher & 64 & 20/10 & 15.6M \\
& Student & 64 & 6/10 & 5.2M \\
\midrule
\midrule
\multirow{2}{*}{NAFNet} & Teacher & 32 & 36/- & 17.1M \\
& Student & 16 & 22/- & 2.7M \\
\midrule
\multirow{2}{*}{Restormer} & Teacher & 48 & 44/- & 26.1M \\
& Student & 24 & 22/- & 3.8M \\
\bottomrule
\end{tabular}
\caption{The specifications of teacher and student models.}
\label{tab:config}
\end{table}

\subsubsection{Datasets and Evaluation} 
For image super-resolution, DCKD is trained using 800 images from DIV2K~\cite{DIV2K} and evaluated on four benchmark datasets. For image deblurring, the models are trained and tested both on GoPro dataset~\cite{GoPro}. For image deraining, we train DCKD on 13,712 clean-rainy image pairs collected from multiple datasets~\cite{Test2800, Rain100H/L, Test100, Rain12} and evaluate it on Test100~\cite{Test100}, Rain100H~\cite{Rain100H/L}, Rain100L~\cite{Rain100H/L}, Test2800~\cite{Test2800}, and Test1200~\cite{Test1200}. We employ the PSNR and SSIM~\cite{SSIM} metrics to evaluate the restoration performance. For image super-resolution and image deraining tasks, the metrics are computed on the Y channel in the YCbCr color space. For image deblurring, PSNR and SSIM are evaluated in the RGB color space.

\subsubsection{Implementation Details}
For image super-resolution, the input is randomly cropped into $48 \times 48$ patches and augmented by random horizontal and vertical flips and rotations. All the models are trained using ADAM optimizer~\cite{kingma2014adam} with $\beta_1=0.9$, $\beta_2=0.99$, and $\epsilon=10^{-8}$. The training batch size is set to 16 with a total of $2.5 \times 10^5$ iterations. The initial learning rate is set to $10^{-4}$ and is decayed by a factor of 10 at every $10^5$ update. DCKD is implemented by PyTorch using 4 NVIDIA V100 GPUs. For other image restoration tasks, we strictly adhere to the original training configurations of each teacher model. More training configurations for other tasks are presented in the Appendix\footnote{https://arxiv.org/abs/2412.08939}.

\begin{table*}[ht]
\small
\centering
\setlength{\tabcolsep}{0.25mm}
\begin{tabular}{c|c|c|c|c|c|c|c|c|c|c|c}
\toprule
\multirow{2}{*}{Scale} & \multirow{2}{*}{Method} & \multirow{2}{*}{Model} & Set5 & Set14 & BSD100 & Urban100 & \multirow{2}{*}{Model} & Set5 & Set14 & BSD100 & Urban100 \\
& & & PSNR/SSIM & PSNR/SSIM & PSNR/SSIM & PSNR/SSIM & & PSNR/SSIM & PSNR/SSIM & PSNR/SSIM & PSNR/SSIM \\
\midrule
\multirow{6}{*}{$\times$2} & Teacher & \multirow{19}{*}{SwinIR} & 38.36/0.9620 & 34.14/0.9227 & 32.45/0.9030 & 33.40/0.9394 & \multirow{19}{*}{RCAN} & 38.27/0.9614 & 34.13/0.9216 & 32.41/0.9027 & 33.34/0.9384 \\
& Scratch & & 38.00/0.9607 & 33.56/0.9178 & 32.19/0.9000 & 32.05/0.9279 & & 38.13/0.9610 & 33.78/0.9194 & 32.26/0.9007 & 32.63/0.9327 \\
& Logits & & 38.04/0.9608 & 33.61/0.9184 & 32.22/0.9003 & 32.09/0.9282 & & 38.17/0.9611 & 33.83/0.9197 & 32.29/0.9010 & 32.67/0.9329 \\
& FAKD & & 38.03/0.9608 & 33.63/0.9182 & 32.21/0.9001 & 32.06/0.9279 & & 38.17/0.9612 & 33.83/0.9199 & 32.29/0.9011 & 32.65/0.9330 \\
& MiPKD & & \underline{38.14}/\underline{0.9611} & \underline{33.76}/\underline{0.9194} & \underline{32.29}/\underline{0.9011} & \underline{32.46}/\underline{0.9313} & & \underline{38.21}/\underline{0.9613} & \underline{33.92}/\underline{0.9203} & \underline{32.32}/\underline{0.9015} & \underline{32.83}/\underline{0.9344} \\
& \textbf{DCKD} & & \textbf{38.17}/\textbf{0.9613} & \textbf{33.85}/\textbf{0.9205} & \textbf{32.31}/\textbf{0.9017} & \textbf{32.59}/\textbf{0.9330} & & \textbf{38.25}/\textbf{0.9615} & \textbf{34.01}/\textbf{0.9213} & \textbf{32.35}/\textbf{0.9019} & \textbf{32.92}/\textbf{0.9351} \\
\cmidrule{1-2}\cmidrule{4-7}\cmidrule{9-12}
\multirow{6}{*}{$\times$3} & Teacher & & 34.89/0.9312 & 30.77/0.8503 & 29.37/0.8124 & 29.29/0.8744 & & 34.74/0.9299 & 30.65/0.8482 & 29.32/0.8111 & 29.09/0.8702 \\
& Scratch & & 34.41/0.9273 & 30.43/0.8437 & 29.12/0.8062 & 28.20/0.8537 & & 34.61/0.9288 & 30.45/0.8444 & 29.18/0.8074 & 28.59/0.8610 \\
& Logits & & 34.44/0.9275 & 30.45/0.8443 & 29.14/0.8066 & 28.23/0.8545 & & 34.61/0.9291 & 30.47/0.8447 & 29.21/0.8080 & 28.62/0.8612 \\
& FAKD & & 34.42/0.9273 & 30.42/0.8437 & 29.12/0.8062 & 28.18/0.8533 & & 34.63/0.9290 & 30.51/0.8453 & 29.21/0.8079 & 28.62/0.8612 \\
& MiPKD & & \underline{34.53}/\underline{0.9283} & \underline{30.52}/\underline{0.8456} & \underline{29.19}/\underline{0.8079} & \underline{28.47}/\underline{0.8591} & & \underline{34.72}/\underline{0.9296} & \underline{30.55}/\underline{0.8458} & \underline{29.25}/\underline{0.8087} & \underline{28.76}/\underline{0.8640} \\
& \textbf{DCKD} & & \textbf{34.59}/\textbf{0.9291} & \textbf{30.57}/\textbf{0.8475} & \textbf{29.22}/\textbf{0.8093} & \textbf{28.63}/\textbf{0.8633} & & \textbf{34.74}/\textbf{0.9299} & \textbf{30.60}/\textbf{0.8472} & \textbf{29.29}/\textbf{0.8099} & \textbf{28.87}/\textbf{0.8662} \\
\cmidrule{1-2}\cmidrule{4-7}\cmidrule{9-12}
\multirow{7}{*}{$\times$4} & Teacher & & 32.72/0.9021 & 28.94/0.7914 & 27.83/0.7459 & 27.07/0.8164 & & 32.63/0.9002 & 28.87/0.7889 & 27.77/0.7436 & 26.82/0.8087 \\
& Scratch & & 32.31/0.8955 & 28.67/0.7833 & 27.61/0.7379 & 26.15/0.7884 & & 32.38/0.8971 & 28.69/0.7842 & 27.63/0.7379 & 26.36/0.7947 \\
& Logits & & 32.27/0.8954 & 28.67/0.7833 & 27.62/0.7380 & 26.15/0.7887 & & 32.45/0.8980 & 28.76/0.7860 & 27.67/0.7400 & 26.49/0.7982 \\
& FAKD & & 32.22/0.8950 & 28.65/0.7831 &  27.61/0.7380 & 26.09/0.7870 & & 32.46/0.8980 & 28.77/0.7860 & 27.68/0.7400 & 26.50/0.7980 \\
& MiPKD & & 32.39/0.8971 & 28.76/0.7854 & 27.68/0.7403 & 26.37/0.7956 & & 32.46/0.8982 & 28.77/0.7860 & 27.69/0.7402 & 26.55/0.7998 \\
& \textbf{DCKD} & & \underline{32.49}/\underline{0.8991} & \underline{28.82}/\underline{0.7877} & \underline{27.72}/\underline{0.7422} & \underline{26.53}/\underline{0.8007} & & \underline{32.56}/\underline{0.8995} & \underline{28.82}/\underline{0.7877} & \underline{27.73}/\underline{0.7423} & \underline{26.69}/\underline{0.8041} \\
& $\mathrm{\textbf{DCKD}}^{*}$ & & \textbf{32.51}/\textbf{0.8992} & \textbf{28.88}/\textbf{0.7890} & \textbf{27.74}/\textbf{0.7430} & \textbf{26.62}/\textbf{0.8032} & & \textbf{32.58}/\textbf{0.8996} & \textbf{28.86}/\textbf{0.7885} & \textbf{27.74}/\textbf{0.7425} & \textbf{26.74}/\textbf{0.8054} \\
\bottomrule
\end{tabular}
\caption{Quantitative comparison on the benchmark datasets for image super-resolution. The best and second-best performances are highlighted in bold and underlined, respectively. The FAKD results on SwinIR are from our reproduction experiments.}
\label{tab:sr} 
\end{table*}

\begin{table}[ht]
\small
\centering
\begin{tabular}{c|c|c|c}
\toprule
\multirow{2}{*}{Model} & \multirow{2}{*}{Method} & GoPro & \multirow{2}{*}{\#Params} \\
& & PSNR/SSIM &  \\
\midrule
MT-RNN & - & 31.15/0.9450 & 2.6M \\
DMPHN & - & 31.20/0.9400 & 21.7M \\
\midrule
\multirow{4}{*}{NAFNet} & Teacher & 32.87/0.9606 & 17.1M \\
& Scratch & 31.17/0.9457 & 2.7M \\
& Logits & 31.26/0.9464 & 2.7M \\
& \textbf{DCKD} & \textbf{31.43}/\textbf{0.9487} & 2.7M \\
\cmidrule{1-4}
\multirow{4}{*}{Restormer} & Teacher & 32.92/0.9610 & 26.1M \\
& Scratch & 31.57/0.9497 & 3.8M \\
& Logits & 31.61/0.9501 & 3.8M \\
& \textbf{DCKD} & \textbf{31.78}/\textbf{0.9521} & 3.8M \\
\bottomrule
\end{tabular}
\caption{Quantitative comparison for image deblurring.}
\label{tab:deblur}
\end{table}

\begin{figure}[ht]
    \centering
    \includegraphics[width=\columnwidth]{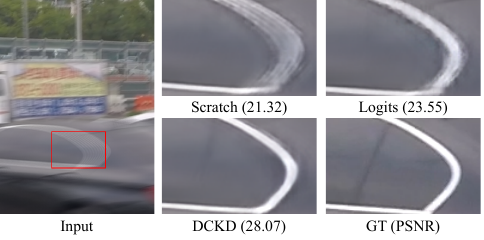}
    \caption{Visual comparison for image deblurring.}
    \label{fig:deblur}
\end{figure}

\subsection{Results and Comparison}

\subsubsection{Image Super-Resolution}
We compare our framework with the representative KD methods: train from scratch, Logits~\cite{hinton2015distilling}, FAKD~\cite{he2020fakd}, and MiPKD~\cite{li2024knowledge}, on $\times2$, $\times3$, and $\times4$ super-resolving scales.

The quantitative results for SwinIR and RCAN are presented in Tab.~\ref{tab:sr}. Existing KD methods provide limited improvement for the student models, and in some cases, they even lead to worse performance compared to models trained without KD. For example, using FAKD for distillation on Urban100 results in worse performance than training the model from scratch on SwinIR. DCKD is effective for both Transformer-based and CNN-based architectures, significantly outperforming existing KD methods by more than 0.1dB across all three scales on Urban100. We deliberately use the most straightforward approach to demonstrate DCKD, showing that dynamic lower-bound constraints can yield strong results even without improving upper-bound constraints. To demonstrate that DCKD can be combined with methods that optimize upper-bound constraints, we incorporate DUKD~\cite{zhang2023data} into the DCKD framework to enhance upper-bound constraints, resulting in DCKD*. As we can see, the proposed DCKD can be combined with existing KD methods that optimize the upper bound, further significantly enhancing performance. DCKD* significantly outperforms the SOTA method MiPKD by 0.25dB on SwinIR and 0.19dB on RCAN at $\times 4$.

\begin{figure*}[ht]
    \centering
    \includegraphics[width=0.95\textwidth]{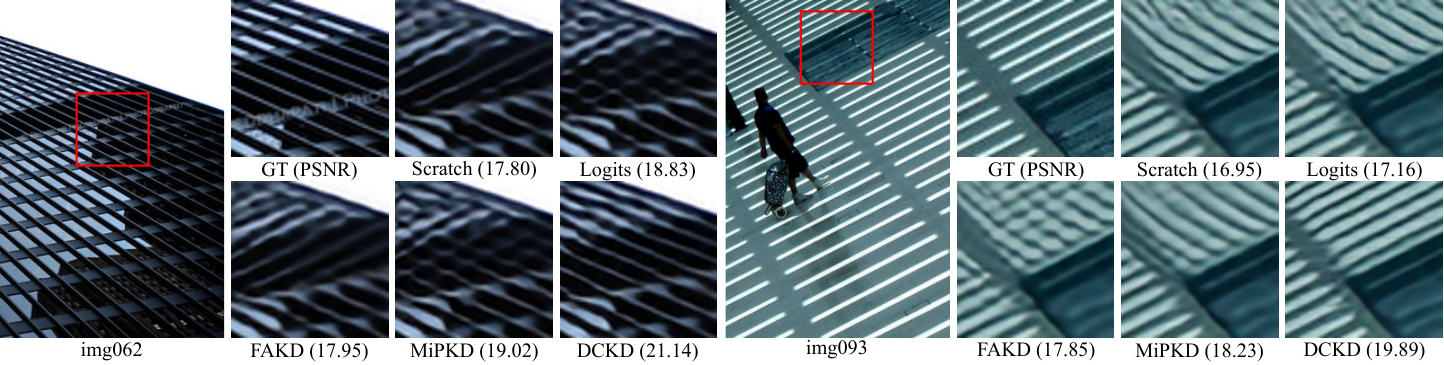}
    \caption{The visual comparison of distilling SwinIR on Urban100 for $\times4$ SR.}
    \label{fig:swinir}
\end{figure*}

\begin{figure*}[ht]
    \centering
    \includegraphics[width=0.95\textwidth]{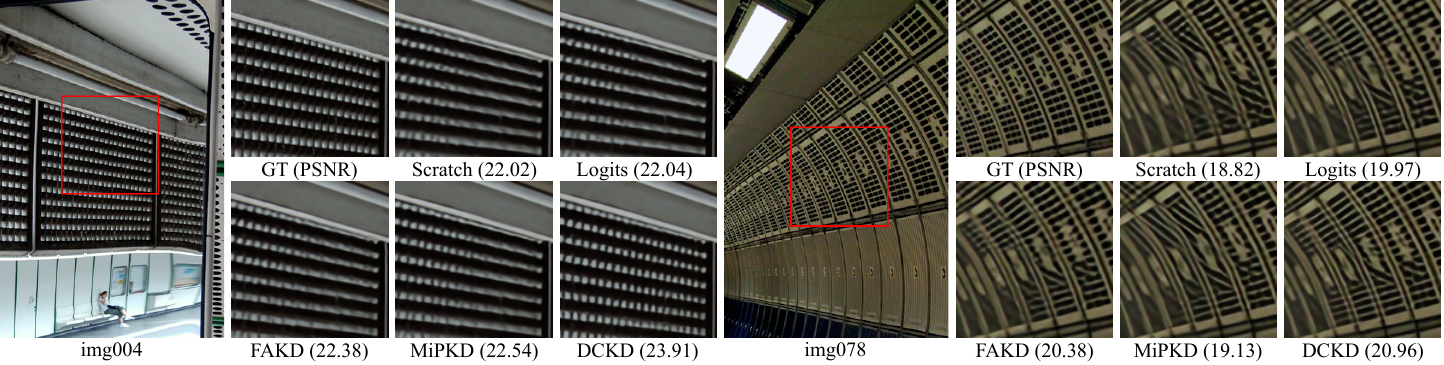}
    \caption{The visual comparison of distilling RCAN on Urban100 for $\times4$ SR.}
    \label{fig:rcan}
\end{figure*}

Fig.~\ref{fig:swinir} and Fig.~\ref{fig:rcan} present challenging visual examples for Transformer and CNN backbones, respectively. Compared to existing KD methods, our approach enables the student models to better capture structural textures, such as more accurately reconstructing sidewalk lines and building structures. More visual comparisons for various examples and models are presented in the Appendix.

\begin{table*}[!ht]
\small
\centering
\begin{tabular}{c|c|ccccc|c}
\toprule
\multirow{2}{*}{Model} & \multirow{2}{*}{Method} & Test100 & Rain100H & Rain100L & Test2800 & Test1200 & \multirow{2}{*}{\#Params} \\
& & PSNR/SSIM & PSNR/SSIM & PSNR/SSIM & PSNR/SSIM & PSNR/SSIM & \\
\midrule
MPRNet & - & 30.27/0.8970 & 30.41/0.8900 & 36.40/0.9650 & 33.64/0.9380 & 32.91/0.9160 & 20.1M \\
SPAIR & - & 30.35/0.9090 & 30.95/0.8920 & 36.93/0.9690 & 33.34/0.9360 & 33.04/0.9220 & - \\
\midrule
\multirow{4}{*}{Restormer} & Teacher & 32.02/0.9237 & 31.48/0.9054 & 39.08/0.9785 & 34.21/0.9449 & 33.22/0.9270 & 26.1M \\
& Scratch & 31.01/0.9122 & 30.51/0.8932 & 37.47/0.9714 & 33.78/0.9396 & 33.67/0.9295 & 3.8M \\
& Logits & 31.04/0.9143 & 30.48/0.8915 & 37.17/0.9712 & 33.81/0.9399 & 33.78/0.9310 & 3.8M \\
& \textbf{DCKD} & \textbf{31.08}/\textbf{0.9167} & \textbf{30.54}/\textbf{0.8969} & \textbf{38.02}/\textbf{0.9762} & \textbf{33.91}/\textbf{0.9411} & \textbf{33.95}/\textbf{0.9326} & 3.8M \\
\bottomrule
\end{tabular}
\caption{Quantitative comparison on the benchmark datasets for image deraining.}
\label{tab:derain}
\end{table*}

\subsubsection{Image Deblurring}
Tab.~\ref{tab:deblur} provides a quantitative comparison on GoPro dataset. Our method demonstrates consistent effectiveness across both CNN-based NAFNet and Transformer-based Restormer. Compared to the Logits KD, our DCKD achieves 0.17dB improvement on different backbones. Moreover, with comparable parameters, the student model of NAFNet significantly outperforms MT-RNN~\cite{park2020multi} by 0.28dB. Fig.~\ref{fig:deblur} illustrates the deblurring visualization results. DCKD restores the clearest window outlines, significantly enhancing the deblurring capability of the student model.

\begin{figure}[ht]
    \centering
    \includegraphics[width=\columnwidth]{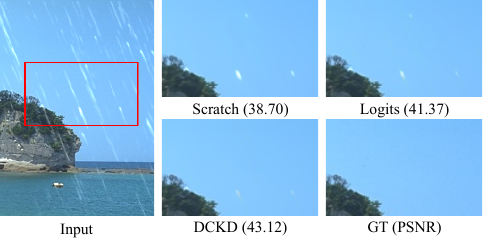}
    \caption{Visual comparison for image deraining.}
    \label{fig:derain}
\end{figure}

\subsubsection{Image Deraining}
Tab.~\ref{tab:derain} shows the performance of various methods on several benchmark datasets for image deraining. With only 18.9\% of MPRNet's~\cite{zamir2021multi} parameters, DCKD significantly surpasses it by 1.8dB on Rain100L dataset. Additionally, compared to other distillation methods, DCKD outperforms the Logits KD by 0.55dB on Rain100L. Fig.~\ref{fig:derain} provides a visual comparison of derained images. DCKD further enhances the student's ability to remove rain streaks compared to the logits distillation method.

\begin{table}[t]
\small
\centering
\begin{tabular}{cc|c|c}
\toprule
 \multirow{2}{*}{DCR} & \multirow{2}{*}{DMM} & Set14 & Urban100 \\
 & & PSNR/SSIM & PSNR/SSIM \\
\midrule
\ding{55} & \ding{55} & 33.83/0.9197 & 32.67/0.9329 \\
 \ding{51} & \ding{55} & 33.98/0.9208 & 32.83/0.9346 \\
\ding{55} & \ding{51} & 33.92/0.9205 & 32.81/0.9343 \\
 \ding{51} & \ding{51} & 34.01/0.9213 & 32.92/0.9351 \\
\bottomrule
\end{tabular}
\caption{Ablation study on components of our framework.}
\label{tab:components}
\end{table}

\begin{table}[t]
\small
\centering
\begin{tabular}{c|c|c}
\toprule
\multirow{2}{*}{Degradation Type} & Set14 & Urban100 \\
& PSNR/SSIM & PSNR/SSIM \\
\midrule
Random Blur & 34.01/0.9212 & 32.89/0.9352 \\
Random Noise & 34.01/0.9213 & 32.92/0.9351 \\
Random Resize & 33.99/0.9212 & 32.90/0.9354 \\
Random Mix & 34.00/0.9212 & 32.87/0.9353 \\
\bottomrule
\end{tabular}
\caption{Ablation study on the degradation module.}
\label{tab:degradation}
\end{table}

\begin{table}[t]
\small
\centering
\begin{tabular}{c|ccccc}
\toprule
$\lambda_{dcl}$ & 0.01 & 0.1 & 1.0 \\
\midrule
PSNR/SSIM & 32.87/0.9346 & 32.92/0.9351 & 32.81/0.9350 \\
\midrule
$\lambda_{ce}$ & 0.0001 & 0.001 & 0.01 \\
\midrule
PSNR/SSIM & 32.84/0.9345 & 32.92/0.9351 & 32.82/0.9341 \\
\bottomrule
\end{tabular}
\caption{Ablation study on the balancing weights.}
\label{tab:weights}
\end{table}

\begin{figure}[t]
    \centering
    \includegraphics[width=\columnwidth]{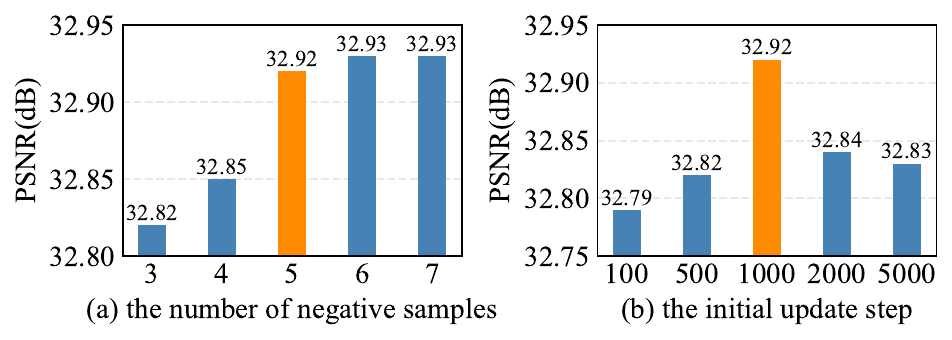}
    \caption{Ablation studies on the number of negative samples and the initial update step.}
    \label{fig:ablation}
\end{figure}

\subsection{Ablation Study}
For ablation experiments, we train DCKD on RCAN for the SR task with a scaling factor of $\times2$. We then validate the results on Set14 and Urban100 datasets.

\subsubsection{Components of the Proposed Framework}
As shown in Tab.~\ref{tab:components}, we first conduct an ablation study on the two main modules of DCKD. The results indicate that our DCR and DMM outperform the baseline by 0.16dB and 0.14dB in PSNR on Urban100, respectively. This demonstrates the effectiveness of the proposed modules. Furthermore, the combination of DCR and DMM further enhances model performance, achieving PSNR improvements of 0.18dB on Set14 and 0.25dB on Urban100 compared to the baseline.

\subsubsection{Impact of the Degradation Module}
We conduct an ablation study on the impact of the degradation module in the Dynamic Negative Sample Generator (DNSG). Following Real-ESRGAN~\cite{wang2021real}, the degradation module is simple to implement by adding the Gaussian blur, Gaussian noise, resize operation (i.e., downsampling and then upsampling), or mixed degradation. The degradation results are summarized in Tab.~\ref{tab:components} and Tab.~\ref{tab:degradation}. We observe that the addition degradation module (in Tab.~\ref{tab:degradation}) achieves at least 0.06dB PSNR improvement on Urban100, compared to that without degradation (only DMM in Tab.~\ref{tab:components}). Furthermore, the degradation with random noise achieves the highest PSNR of 32.92dB, which outperforms randomly mixed degradation by 0.05dB PSNR.

\subsubsection{Impact of the Balancing Weights}
We investigate the impact of the balancing coefficients $\lambda_{dcl}$ and $\lambda_{ce}$ in Equ.~\ref{equ:all}, as shown in Tab.~\ref{tab:weights}. We find that excessively large or small values for these coefficients negatively affect the outcomes. The experiments indicate that the model achieves optimal results when $\lambda_{dcl}$ is set to 0.1 and $\lambda_{ce}$ to 0.001. Given the broad applicability of our method to various image restoration tasks, we adopt $\lambda_{dcl}=0.1$ and $\lambda_{ce}=0.001$ as the default settings across different tasks.

\subsubsection{Impact of the Number of Negative Samples}
The impact of the number of negative samples is reported in Fig.~\ref{fig:ablation} (a). The results indicate that as the number of negative samples increases, the performance improves consistently. However, when the number of negative samples exceeds 5, the performance gains diminish while significantly increasing training time and memory costs. Therefore, the number of negative samples is set to 5, which achieves the best trade-off between PSNR and training time.

\subsubsection{Impact of the Initial Update Step}
In Fig.~\ref{fig:ablation} (b), we investigate the impact of the initial update step for the history model within the dynamic negative sample generator. The experimental results show that when using a smaller step to update the historical model, the quality of the negative samples becomes very close to, or even surpasses, that of the anchor points, leading to instability in the solution space and a decline in performance. Conversely, when using a larger step, the negative sample quality deteriorates, weakening the lower bound constraint. When the initial update step is set to 1000 achieve the best performance.

\section{Conclusion}
In this work, we propose a dynamic contrastive knowledge distillation framework for image restoration, named DCKD, which consists of the Dynamic Contrastive Regularization (DCR) and the Distribution Mapping Module (DMM). Most previous knowledge distillation methods utilize a fixed solution space, causing the lower bound constraints to weaken gradually during training. DCR constructs a dynamic solution space based on the student's learning state to enhance the lower-bound constraints. DMM introduces pixel-level category information to knowledge distillation for low-level vision tasks for the first time. Experiments on image super-resolution, image deblurring, and image deraining tasks validate that the proposed DCKD achieves state-of-the-art results on various benchmark datasets, both quantitatively and visually.

\begin{table*}[ht]
\small
\centering
\begin{tabular}{c|c|c|cccc}
\toprule
\multirow{2}{*}{Model} & \multirow{2}{*}{Scale} & \multirow{2}{*}{Method} & Set14 & BSD100 & Urban100 & Manga109 \\
& & & PSNR/SSIM & PSNR/SSIM & PSNR/SSIM & PSNR/SSIM \\
\midrule
\multirow{3}{*}{SwinIR-light} & \multirow{3}{*}{$\times$4} & - & 28.77/0.7858 & 27.69/0.7406 & 26.47/0.7980 & 30.92/0.9151 \\
& & MCLIR & 28.85/0.7874 & 27.72/0.7414 & 26.57/0.8010 & 31.04/0.9158 \\
& & \textbf{DCKD} & \textbf{28.88}/\textbf{0.7891} & \textbf{27.75}/\textbf{0.7432} & \textbf{26.64}/\textbf{0.8039} & \textbf{31.20}/\textbf{0.9180} \\
\bottomrule
\end{tabular}
\caption{Quantitative comparison DCKD with MCLIR~\cite{wu2024learning} on the benchmark datasets for SR task.}
\label{tab:cl}
\end{table*}

\begin{figure*}[!ht]
    \centering
    \includegraphics[width=0.95\textwidth]{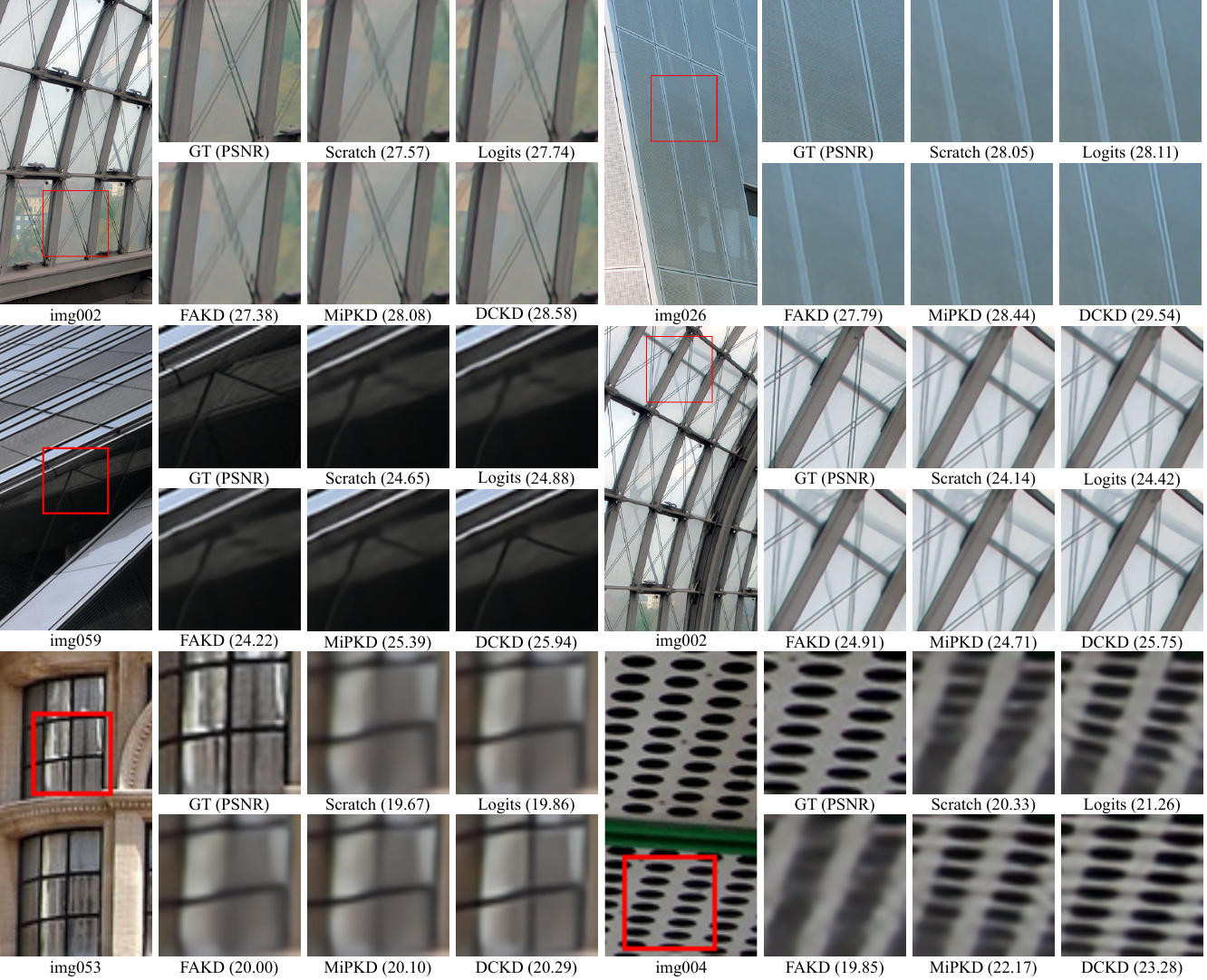}
    \caption{More visual comparison of image super-resolution at $\times4$.}
    \label{fig:sr}
\end{figure*}

\begin{figure*}[ht]
    \centering
    \includegraphics[width=0.95\textwidth]{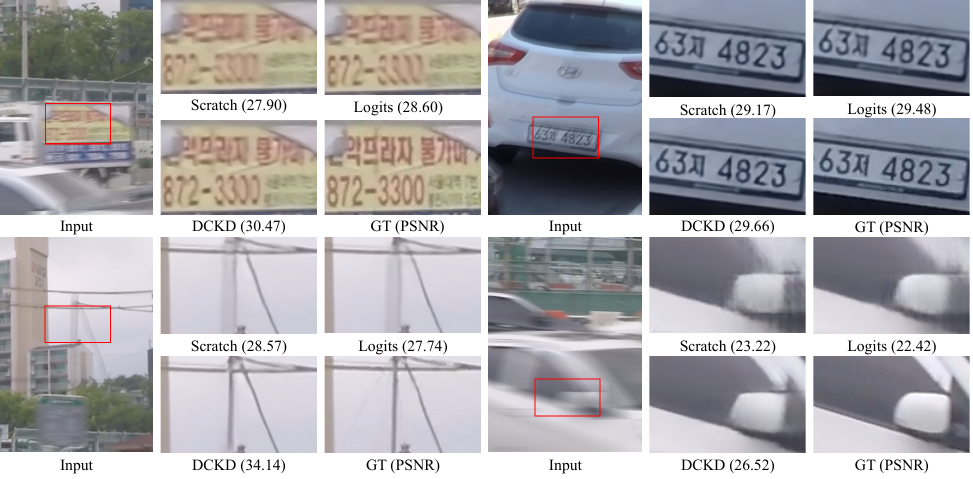}
    \caption{More visual comparison of image deblurring.}
    \label{fig:blur}
\end{figure*}

\begin{figure*}[ht]
    \centering
    \includegraphics[width=0.95\textwidth]{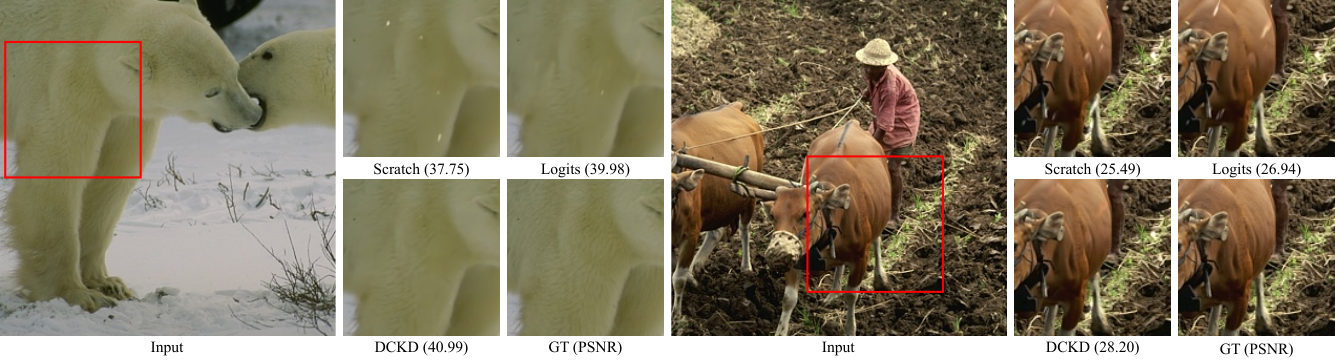}
    \caption{More visual comparison of image deraining.}
    \label{fig:rain}
\end{figure*}

\section{Acknowledgments}
This work is supported by the National Natural Science Foundation of China (NO. 62102151), the Open Research Fund of Key Laboratory of Advanced Theory and Application in Statistics and Data Science, Ministry of Education (KLATASDS2305), the Fundamental Research Funds for the Central Universities. 

\section{Appendix}

\subsection{Implementation Details}

\subsubsection{The details of image encoder.}
For the latent features, we extract the features from five different layers of the pre-trained VQGAN~\cite{esser2021taming}, while with the corresponding coefficients $\lambda_i,i=1,\cdots5$ to $\frac{1}{32},\frac{1}{16},\frac{1}{8},\frac{1}{4}$ and 1, respectively. We set the attenuation rate $\alpha$ to 0.1.

\subsubsection{The details of different backbones.}
For NAFNet~\cite{chen2022simple}, we train the model with AdamW optimizer ($\beta_1=0.9$, $\beta_2=0.9$, weight deacy = $10^{-3}$) for total $2 \times 10^5$ iterations with the initial learning rate $10^{-3}$ gradually reduced to $10^{-7}$ with the cosine annealing schedule~\cite{loshchilov2016sgdr}. The training patch size is 256 × 256 and the batch size is 32.

For Restormer~\cite{zamir2022restormer}, we train the model with AdamW optimizer ($\beta_1=0.9$, $\beta_2=0.999$, weight deacy = $10^{-4}$) for $3 \times 10^5$ iterations with the initial learning rate $3 \times 10^{-4}$ gradually reduced to $1 \times 10^{-6}$ with the cosine annealing~\cite{loshchilov2016sgdr}. We start training with patch size $128 \times 128$ and batch size 64 for progressive learning. The patch size and batch size pairs are updated to [$(160^2,40), (192^2,32), (256^2,16), (320^2,8), (384^2,8)$] at iterations [92K, 156K, 204K, 240K, 276K]. For data augmentation, we use horizontal and vertical flips.

\subsection{Comparison of DCKD with Contrastive Learning}
The quantitative comparison of DCKD with contrastive learning on the benchmark datasets for the SR task is reported in Tab.~\ref{tab:cl}. DCKD significantly surpasses state-of-the-art method MCLIR~\cite{wu2024learning} 0.16dB on Manga109. This shows that DCKD provides more comprehensive lower-bound constraints, ensuring that each negative sample exerts a consistent force on the anchor sample. Additionally, our method requires only a single negative sample model to generate an arbitrary number of negative samples, significantly reducing computational overhead.

\subsection{More Visual Comparison}
We provide more visual comparisons. Fig.~\ref{fig:sr} presents some challenging images from the x4 super-resolution task. As observed, previous distillation methods often result in blurred artifacts and a struggle to effectively restore high-frequency details such as lines and architectural structures. For example, in img002 and img026, previous methods produce significant artifacts between lines or blur the gaps between them. In contrast, our DCKD not only removes these artifacts but also clearly separates the lines. Overall, our method enhances the student model's ability to handle high-frequency details and accurately restore them.

Further visual comparisons for deblurring and deraining tasks are shown in Fig.~\ref{fig:blur} and Fig.~\ref{fig:rain}, where our DCKD effectively removes motion blur and rain, with the restored image quality closely matching that of the ground truth (GT).


\end{document}